\documentclass[10pt,twocolumn,letterpaper]{article}
\usepackage{cvpr}

\usepackage{cite}
\usepackage[inline]{enumitem}
\usepackage{subfig}
\usepackage{booktabs} 
\newcommand{\ra}[1]{\renewcommand{\arraystretch}{#1}}
\usepackage{microtype}   
\usepackage{ccicons}
\usepackage{multirow}

\usepackage[breaklinks=true,bookmarks=false]{hyperref}
\cvprfinalcopy
\def\cvprPaperID{****} 

\usepackage{graphicx}
\usepackage{amsmath,amssymb} 
\usepackage{color}
\usepackage{bbm}

\begin{document}

%
%
%
%

\title{DeepDiary: Automatic Caption Generation for \\ Lifelogging Image Streams}

\author{
Chenyou Fan \ \ \ \ \ \ \ \ \ David J. Crandall\\
School of Informatics and Computing, Indiana University\\
Bloomington, Indiana USA\\
{\tt\small \{fan6,djcran\}@indiana.edu}
}

\maketitle

\begin{abstract}
Lifelogging cameras capture everyday life from a first-person
perspective, but generate so much data that it is hard for users to
browse and organize their image collections effectively.  In this
paper, we propose to use automatic image captioning algorithms to
generate  textual representations of these collections.  We
develop and explore novel techniques based on deep learning to generate
captions for both individual images and image streams, using temporal consistency
constraints to create summaries that are both more compact and less
noisy.  
We evaluate our techniques with
quantitative and qualitative results, and apply captioning to an image
retrieval application for finding potentially private images.
Our results suggest that our automatic captioning algorithms, while imperfect,
may work well enough to help users manage lifelogging photo collections.\footnote{This is an expanded preprint of a paper appearing at the ECCV International Workshop on Egocentric Perception, Interaction, and Computing~\cite{deepdiary2016eccvw}.}
\end{abstract}

\section{Introduction}

Wearable cameras that capture first-person views of people's daily
lives have recently become affordable, lightweight, and practical,
after many years of being explored only in the research
community~\cite{hodges2006sensecam,mann2002sousveillance,azuma2001recent}.
These new devices come in various types and styles, from the GoPro,
which is marketed for recording high-quality video of sports and other
adventures, to Google Glass, which is a heads-up display interface for
smartphones but includes a camera, to Narrative Clip
and Autographer, which capture ``lifelogs'' by
automatically taking photos throughout one's day (e.g., every 30 seconds). 
These devices, and others like them, are being used for a variety
of applications, from documenting police officers' interactions with
the public~\cite{policecams}, to studying people's activities at a
fine grain resolution for psychological
studies~\cite{kalnikaite2010now,doherty2011passively}, to keeping
visual diaries of people's lives for promoting
health~\cite{moghimi2014sedentary} or 
just for personal use~\cite{gurrin2008examination,doherty2012experiences}. 
No matter the purpose, however, all of these devices can record huge
amounts of imagery, which makes it difficult for users to organize and
browse their image data.  

In this paper, 
we attempt to produce automatic textual \textit{narrations} or
captions of a visual lifelog.  We believe that describing lifelogs
with sentences is most natural for the average user, and allows for
interesting applications like generating automatic textual diaries of
the ``story'' of someone's day based on their lifelogging photos.  We
take advantage of recent breakthroughs in image captioning using deep
learning that have shown impressive results for consumer-style images
from social media~\cite{karpathy2014deep,karpathy2014fragment}, and
evaluate their performance on the novel domain of first-person images
(which are significantly more challenging due to substantial noise,
blurring, poor composition, etc.).  We also propose a new 
strategy to try to encourage diversity in the sentences,
which we found to be particularly useful in describing lifelogging images
from different perspectives.

Of course, lifelogging photo streams are highly redundant since
wearable cameras indiscriminately capture thousands of photos per
day. Instead of simply captioning individual images, we also consider
the novel problem of jointly captioning lifelogging streams,
i.e. generating captions for temporally-contiguous groups of photos
corresponding to coherent activities or scene types.  Not only does
this produce a more compact and potentially useful organization of a
user's photo collection, but it also could create an
automatically-generated textual ``diary'' of a user's day based only
on their photos. The sentences themselves are also useful to aid in
image retrieval by keyword search, which we illustrate for the
specific application of searching for potentially private images
(e.g. containing keywords like ``bathroom''). 
 Joint caption estimation over multiple images also reduces noise and errors in the
captioning results, since evidence from multiple photos is used to
infer each caption.  
We formulate this joint captioning problem in a
Markov Random Field model and show how to solve it efficiently.


To our knowledge, we are the first to propose image captioning as an important task for lifelogging photos,
as well as the first to apply and evaluate automatic image captioning models in this domain.
To summarize our contributions, we learn and apply deep image captioning models to lifelogging photos, including proposing a novel method for generating photo descriptions with diverse structures and perspectives; propose a novel technique for inferring captions for streams of photos taken over time in order to find and summarize coherent activities and other groups of photos; create an online framework for collecting and annotating lifelogging images, and use it to collect a realistic lifelogging dataset consisting of thousands of photos and thousands of reference sentences and evaluate these techniques on our data, both quantitatively and qualitatively, under different simulated use cases.
%
%


\section{Related Work}



While wearable cameras have been studied for over a
decade in the research community~\cite{hodges2006sensecam,mann2002sousveillance,azuma2001recent}, only recently have they become practical enough for consumers to use on a daily basis.
Recent work has explored using them to aid human memory for
retrospection~\cite{clinch2014lifelogging,
  yoo2013facelog,doherty2012experiences,gurrin2008examination}, to help
students learn~\cite{barreau2006augmenting}, to assist people with visual
impairments~\cite{karim2006exploiting}, and to study people's activities at a
fine grain resolution for psychological
studies~\cite{kalnikaite2010now,doherty2011passively,egohands2015iccv}, among many
other applications.
These wearable camera applications raise a number of challenges.  From
a privacy perspective, for example, 
Denning \etal~\cite{denning2014situ} and Nguyen
\etal~\cite{nguyen2009encountering} study how bystanders react to
wearable cameras, while Hoyle \etal~\cite{hoyle2014privacy} identify
privacy risks to the camera wearers themselves.  From a technical
standpoint, many applications would require automatic techniques to
analyze and organize the vast quantities of images that wearable
cameras collect.   
In the computer vision field, recent work has begun
to study this new style of imagery, which is significantly different
 from photos taken by traditional point-and-shoot cameras.
Specific research topics have included recognizing
objects~\cite{objectavoider,fathi2011learning},
scenes~\cite{furnari2015}, and activities~\cite{fathi2012learning, ryoo2015robot,castro2015activities, ryoo2013first}.  Some
computer vision work has specifically tried to address privacy
concerns, by recognizing photos taken in potentially sensitive places
like bathrooms~\cite{templeman2014placeavoider}, or containing
sensitive objects like computer
monitors~\cite{objectavoider}.  However, these techniques typically
require that classifiers be explicitly trained for each object,
scene type, or activity of interest, which limits their scalability.

Instead of classifying lifelogging images into pre-defined and discrete
categories, we propose to
annotate them with automatically-generated, free-form
image captions, inspired by recent progress in deep learning.
Convolutional Neural Networks
(CNNs) have recently emerged as powerful models for object
recognition in computer vision~\cite{szegedy2013deep,
  erhan2014scalable, girshick2014rich, krizhevsky2012imagenet},
while 
Recurrent Neural Networks (RNNs) and Long Short-Term Memory (LSTMs) have
been developed for learning models of sequential data, like natural
language sentences~\cite{elman1990finding, graves2013generating}.
The combination of CNNs for recognizing image content and RNNs for
modeling language have recently been shown to  generate surprisingly rich image
descriptions~\cite{mao2014explain, vinyals2014show, karpathy2014deep},
essentially ``translating'' from image features to English
sentences~\cite{karpathy2015visualizing}.

Some closely related work has been done to generate textual descriptions from videos.
Venugopalan
\etal~\cite{venugopalan2015sequence} use image captioning model to
generate video descriptions from a sequence of video frames. Like
previous image captioning papers, their method estimates 
a single sentence for each sequence, while we explicitly generate multiple
diverse sentences and evaluate the image-sentence
matching quality to improve the captions from noisy, poorly-composed lifelogging images. Zhu
\etal~\cite{zhu2015aligning} use neural sentence embedding to model a
sentence-sentence similarity function, and use LSTMs to model
image-sentence similarity  in order to align subtitles of
movies with sentences from the original books. Their main purpose is to
find corresponding movie clips and book paragraphs based on visual and
semantic patterns, whereas ours is to infer novel sentences from new
lifelogging image streams.

\section{Lifelogging Data Collection}

To train and test our techniques, two of the authors wore Narrative
Clip lifelogging cameras over a period of about five months (June-Aug 2015 and Jan-Feb 2016), to create
a repository of 7,716 lifelogging photos.  To facilitate collecting
lifelogging photos and annotations, we built a website which allowed
users to upload  and label photos in a unified framework, using
the Narrative Clip
API.\footnote{https://open-staging.getnarrative.com/api-docs}


We collected textual annotations for training and testing the system
in two different ways. First, the two authors and three of their
friends and family members used the online system to submit sentences for
randomly-selected images, producing {2,683} sentences for {696}
images.  Annotators were asked to produce at least two sentences per
image: one that described the photo from a first-person perspective
(e.g., ``I am eating cereal at the kitchen table.'') and one from a
third-person perspective (e.g. ``A bowl of cereal sits on a kitchen
table.''). We requested sentences from each of these perspectives
because we have observed that some scenes are more naturally described by one perspective or the other. Annotators were welcome
to enter multiple sentences, and each image was viewed by an average
of 1.45 labelers. 

Second,
to generate more diversity in annotators and annotations, we published
{293} images\footnote{We randomly chose 300, but removed 7 that we were not comfortable sharing with the public  (e.g.\ photos of strangers whose permission we were not able to obtain).} on Amazon's Mechanical Turk (AMT), showing each
photo to at least three annotators and, as before, asking each annotator
to give at least one first-person and one third-person sentence.  This
produced a set of {1,813} sentences, or an average of
{6.2} sentences per image. A total of {121} distinct Mechanical
Turk users contributed sentences.  

Finally, we also downloaded COCO~\cite{mscoco}, a popular publicly-available dataset of 80,000 photos
and 400,000 sentences. These images are from Internet and social media sources, and thus are significantly
different than the lifelogging context we consider here, but we hypothesized that this may be useful additional training data
 to augment our smaller lifelogging dataset.



\section{Automatic Lifelog Image Captioning}  
\label{sect:auto_caption}

We now present our technique for using deep learning to automatically
annotate lifelogging images with captions. We first give a brief
review of deep image captioning models, and then show how to
take advantage of \textit{streams} of lifelogging images by estimating
captions jointly across time, which not only helps reduce noise in
captions by enforcing temporal consistency, but also helps
summarize large photo collections with smaller subsets of sentences.

\subsection{Background: Deep networks for captioning}

Automatic image captioning is a difficult task because it requires not
only identifying important objects and actions, but also describing
them in natural language.  However, recent work in deep learning has
demonstrated impressive results in generating image and video
descriptions~\cite{karpathy2014deep,venugopalan2015sequence,zhu2015aligning}.
The basic high-level idea is to learn a common feature space that is
shared by both images and words. Then, given a new image, we generate
sentences that are ``nearby'' in the same feature space.  The encoder
(mapping from image to feature space) is typically a Convolutional
Neural Network (CNN), which abstracts images into a vector of local
and global appearance features. The decoder (mapping from feature
space to words) produces a word vector using a Recurrent Neural Network (RNN), which abstracts out the semantic and syntactic meaning.

For extracting visual features, Convolutional Neural
Networks (CNNs)~\cite{krizhevsky2012imagenet} have become very
popular.  
A typical CNN is much like a classical feed-forward neural
network, except that the connections between early layers are not all
fully-connected, but instead have specially-designed structures with
shared weights that encode operations like convolutions and spatial
pooling across local image regions.  
Modern CNNs are also typically very deep, often with 20 or more layers. While the final output of CNNs differ based on the task (e.g., a class label for image classification), a common trick is to use the output of one of the penultimate layers as a feature vector to represent the visual appearance of an input image.  
These ``deep features'' produced by CNNs have been repeatedly
shown~\cite{razavian2014cnn,girshick2014rich} to outperform
traditional hand-made image features such as
SIFT~\cite{lowe2004distinctive} and HOG~\cite{dalal2005histograms}.

\begin{figure}[t]
\begin{center}
   \includegraphics[width=\columnwidth,trim=0cm 0cm 2cm 0cm,clip]{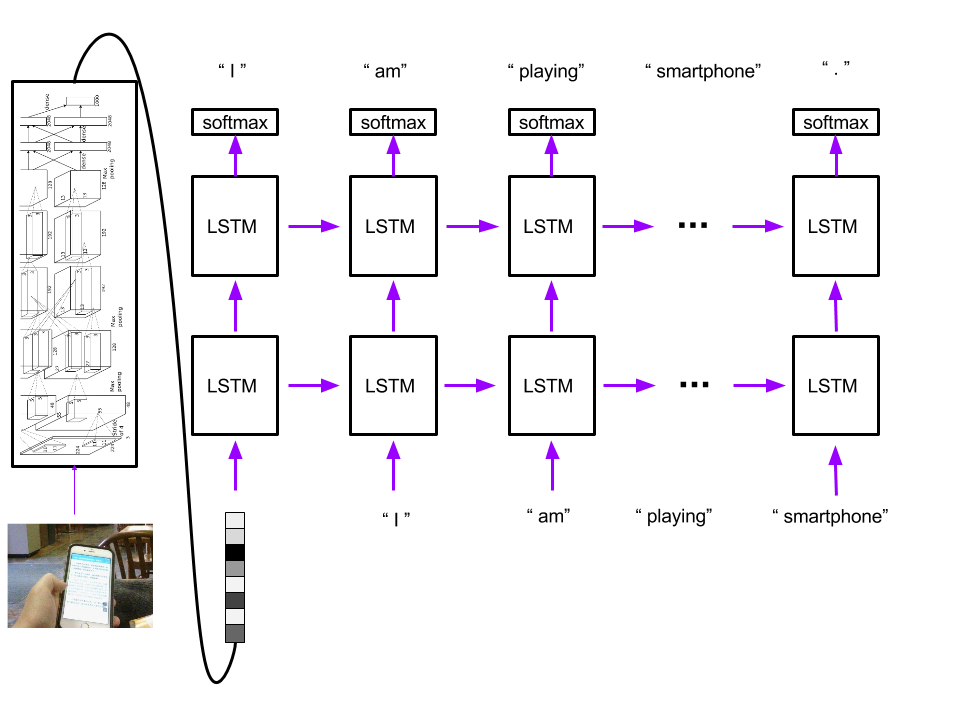}
\end{center}
\vspace{-20pt}
   \caption{Illustration of deep image captioning using a CNN and a
     two-layer LSTM. An image is fed into a CNN to produce a visual
     feature representation, and presented to the LSTM as its initial
     input.  Then each word in a training sentence is presented to the LSTM at
     each step. A softmax layer is attached to the second layer of the
     LSTM to generate predicted sentences and softmax loss. }
\label{fig:lstm_model}		
\end{figure}

For modeling sequences like sentences, Recurrent Neural Networks
(RNNs)~\cite{elman1990finding, graves2013generating}, and specifically
Long Short-Term Memory (LSTM) models~\cite{hochreiter1997long,
  graves2013generating}, have become popular.  
RNNs 
include hidden units that are self-connected (i.e. some of their
inputs are connected to their own outputs), which have the effect of
having ``memory'' that can develop internal representations for the
patterns of input sequences~\cite{elman1990finding}.  LSTMs are a
special form of RNNs that include an array of specially-designed
hidden units called memory blocks, each of which contains three gates
and a memory cell.
 In any given iteration of training, a memory block can choose to read or ignore its input, to remember or forget its current cell value, or to output or suppress the new cell value. 
These two ingredients of CNN and LSTM models are usually combined for
image captioning in the following
way~\cite{vinyals2014show,karpathy2014deep,venugopalan2015sequence}.
The training data consists of a set of images, each with at least one
human-generated reference sentence.  During training, for any given
image $I$, we first generate a corresponding deep visual feature
vector $v_I$ using a CNN.  This vector is then presented as the
initial input to the LSTM model.  Then, the LSTM model is presented
with each word in the training sentence in turn, by inputing the
word vector corresponding to each word to the LSTM.  The output from the hidden unit predicts the next word in the sentence, in particular giving a probability distribution over words in the dictionary. During each step $t$ of training, error is back-propagated from the following step as well as from the softmax (word generation) layer. 
Word vectors as well as weights of hidden units are trained and updated during back-propagation.  
An intuitive way to visualize LSTM is to unroll the hidden states at each time, as shown in
Figure~\ref{fig:lstm_model}. 
As shown in the figure,
in practice we use a two-layer LSTM 
to get better predictions, such that the hidden states
of the first layer serve as input to the second layer and word predictions are
emitted from the output of the second layer. 

At test time, to generate a caption for new image $I'$, we again use a CNN to produce an image feature vector and present it to the LSTM, which then predicts the first word of the sentence based on the visual features.  After that, the best prediction at step $t$ for word $w_t$ is used as the input for step $t+1$.  
In the prediction stage, a forward pass of LSTM generates a full
sentence terminated by a stop word for each input image.  Similar
image captioning models have been discussed in detail in recent
papers~\cite{vinyals2014show,karpathy2014deep,venugopalan2015sequence}. In
section~\ref{diverse_caption}, we discuss in detail how to
generate diverse captions for a single image.

\subsection{Photo Grouping and Activity Summarization}

The techniques in the last section automatically estimate captions for
individual images.  However, lifelogging users do not typically
capture individual images in isolation, but instead collect long
streams of photos taken at regular intervals over time (e.g., every 30
seconds for Narrative Clip).  
This is a significant difference from
most applications of image captioning that have been studied before,
which target isolated images found on the Internet or in social media,
and represents both a challenge and an opportunity. The challenge is
that generating thousands of captions for a day's activities, one for
each photo, could easily overwhelm a user.  
This means that
evidence from multiple images can be combined together to
produce better captions than is possible from observing any single
image, in effect ``smoothing out'' noise in any particular image by
examining the photos taken nearby in time.  These 
sentences could provide more concise summarizations, helping people
find, remember, and organize photos according to broad events instead
of individual moments. 


Suppose we wish to estimate captions for a stream of images $I=(I_1,
I_2, ..., I_K),$ which are sorted in order of increasing timestamps.
We first generate multiple diverse captions for each individual image,
using a technique we describe in the next subsection. We combine all
of these sentences together across images into a large set of
candidates $C$ (with $|C|=d|I|$, where $d$ is the number of diverse
sentences generated per image; we use $d=15$).
We wish to estimate a sequence
of sentences such that each sentence describes its corresponding image
well, but also such that the sentences are relatively consistent across
time. In other words, we want to estimate a sequence of sentences
$S^*=(S^*_1, S^*_2, ..., S^*_K)$ so as to minimize an energy function,
\newcommand{\argmin}{\operatornamewithlimits{argmin}}
\begin{equation} 
S^* = \argmin_{S=(S_1, ..., S_K)} \sum_{i=1}^K \text{Score}(S_i, I_i) +\beta\sum_{j=1}^{K\text{-}1}\mathbbm{1}(S_j, S_{j\text{+}1}), 
\label{eq:mrf}
\end{equation}
where each $S_i \in C$, $\text{Score}(S_i, I_i)$ is a unary cost
function measuring the quality of a given sentence $S_i$ in describing
a single image $I_i$, $\mathbbm{1}(S_a, S_b)$ is a pairwise cost
function that is 0 if $S_a$ and $S_b$ are the same and 1 otherwise,
and $\beta$ is a constant. Intuitively, $\beta$ controls the
degree of temporal smoothing of the model: when $\beta=0$, for
example, the model simply chooses sentences for each image
independently without considering neighboring images in the stream,
whereas when $\beta$ is very large, the model will try to find a
single sentence to describe all of the images in the stream.

Equation~(\ref{eq:mrf}) is a chain-structured Markov Random Field
(MRF) model~\cite{koller}, which means that the optimal sequence of
sentences $S^*$ can be found efficiently using the Viterbi algorithm.
All that remains is to define two key components of the model: (1) a
technique for generating multiple, diverse candidate sentences for
each image, in order to obtain the candidate sentence set $C$, and (2)
the $\text{Score}$ function, which requires a technique for
measuring how well a given sentence describes a given image.
We now describe these two ingredients in turn.

\subsubsection{Generating Diverse Captions} \label{diverse_caption}
Our joint captioning model above  requires a large set of candidate
sentences. 
Many possible sentences can correctly describe any given image, and
thus it is desirable for the automatic image captioning algorithm to
generate multiple sentences that describe the image in multiple ways.
This is especially true for lifelogging images that are often noisy,
poorly composed, and ambiguous, and can be interpreted in different
ways.  Vinyals \etal~\cite{vinyals2014show} use beam search to
generate multiple sentences, by having the LSTM model keep $b$
candidate sentences at each step of sentence generation (where $b$ is
called the beam size).  However, we found that this existing technique
did not work well for lifelogging sentences, because it produced very
homogeneous sentences, even with a high beam size.

\begin{figure*}[t]
\begin{center}
   \includegraphics[width=0.8\textwidth,trim=1.2cm 0.5cm 0.3cm 0.5cm,clip]{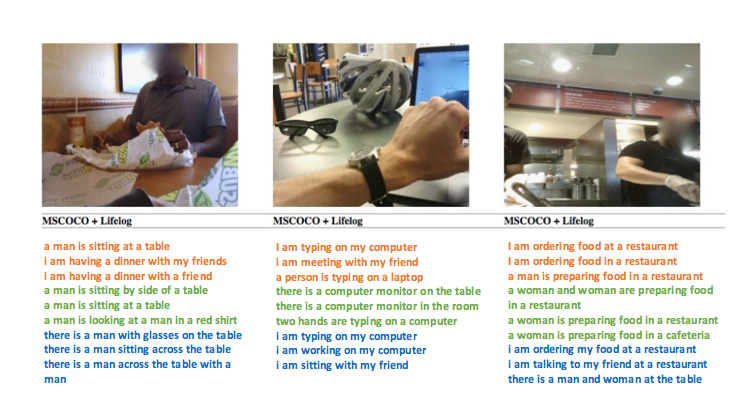}
\end{center}
\vspace{-12pt}
   \caption{Sample captions generated by models pre-trained with COCO and
     fine-tuned with lifelogging dataset. Three different colors show the
     top three predictions produced in three beam searches by applying the
     Diverse M-Solutions technique. Within each beam search, sentences
     tend to have similar structures and describe from similar
     perspective; between consecutive beam searches, structures and
     perspectives tend to be different. }
\label{fig:full_captions}		
\end{figure*}

To encourage greater diversity, we apply the Diverse M-best solutions
technique of Batra \etal~\cite{batra2012diverse}, which was originally
proposed to find multiple high-likelihood solutions in graphical model inference problems.  
We adapt this technique to LSTMs by performing multiple rounds of beam
search. In the first round, we obtain a set of predicted words for each
position in the sentence. In the second round, we add a bias term that
reduces the network activation values of words found in the first beam
search by a constant value.  Intuitively, this decreases the probability
that a word found during the previous beam search being selected again at
the same word position in the sentence. Depending on the degree of diversity needed,
additional rounds of beam search can be conducted, each time penalizing
words that have occurred in any previous round.
In our current implementation, we use three rounds of beam search and
set the beam size to be five, so we generate a total of 15 candidate
sentences for each individual image. The set of all of these sentences
across all images in the photo stream produces the candidate sentence
set $C$ in equation (\ref{eq:mrf}).
Figure~\ref{fig:full_captions}
presents sample automatically-generated results by using three rounds of beam search and a beam size of 3 for illustration purposes.  We see that the technique successfully
injects diversity into the set of estimated captions. Many of the
captions are quite accurate, including ``A man is sitting at a table''
and ``I am having dinner with my friends,'' while others are not
correct (e.g. ``A man is looking at a man in a red shirt''), and
others are nonsensical (``There is a man sitting across the table with
a man''). Nevertheless, the captioning results are overall remarkably
accurate for an automatic image captioning system, reflecting the 
power of deep captioning techniques to successfully model both
image content and sentence generation.

\subsubsection{Image-sentence quality alignment}
\label{sect:alignment}

The joint captioning model in Equation~(\ref{eq:mrf}) also requires a
function $\text{Score}(S_i, I_i)$, which is a measure of how well an
arbitrary sentence $S_i$ describes a given image $I_i$. The difficulty
here is that the LSTM model described above tells us how to generate
sentences for an image, but not how to measure their similarity to a given
image. Doing this requires us to explicitly align certain words
of the sentence to certain regions of an image~-- i.e. determining
which ``part'' of an image generated each word.
Aligning between words of a sentence with regions of an image is a
difficult task.  Common captioning datasets such as COCO~\cite{mscoco}
contain image-level captions, but do not have ground truth data
mapping words to specific objects or image regions.
Karpathy~\etal~\cite{karpathy2014fragment} propose matching each
region with the word with maximum inner product (interpreted as a
similarity measure) across all words in terms of learnable region
vectors and word vectors, and to sum all similarity measures over all
regions as the total score.
They use Regions with CNN
(R-CNN)~\cite{girshick2014rich} to detect image regions and obtain
region feature vectors.  Word vectors are encoded by Bidirectional
LSTM (BLSTM)~\cite{schuster1997bidirectional, graves2013hybrid} (which
is a variant of LSTM that captures contextual information from not
only previous words but also future ones).  They also construct
positive training instances (true image-description pairs) and
negative training instances (images with randomly sampled sentences
from other images) to train this model in an unsupervised fashion
which seeks to maximize the margin between scores of positive training
instances and negative instances. 
We implement their method and train
this image-sentence alignment model on our lifelogging dataset.  To
generate the matching score $\text{Score}(S_i, I_i)$ for
Equation~(\ref{eq:mrf}), we extract region vectors from image $I_i$,
retrieve trained word vectors for words in sentence $S_i$, and sum
similarity measures of regions with best-aligned words.

\subsubsection{Image grouping}
\begin{figure*}[t]
\begin{center}
   \includegraphics[width=1.5\columnwidth]{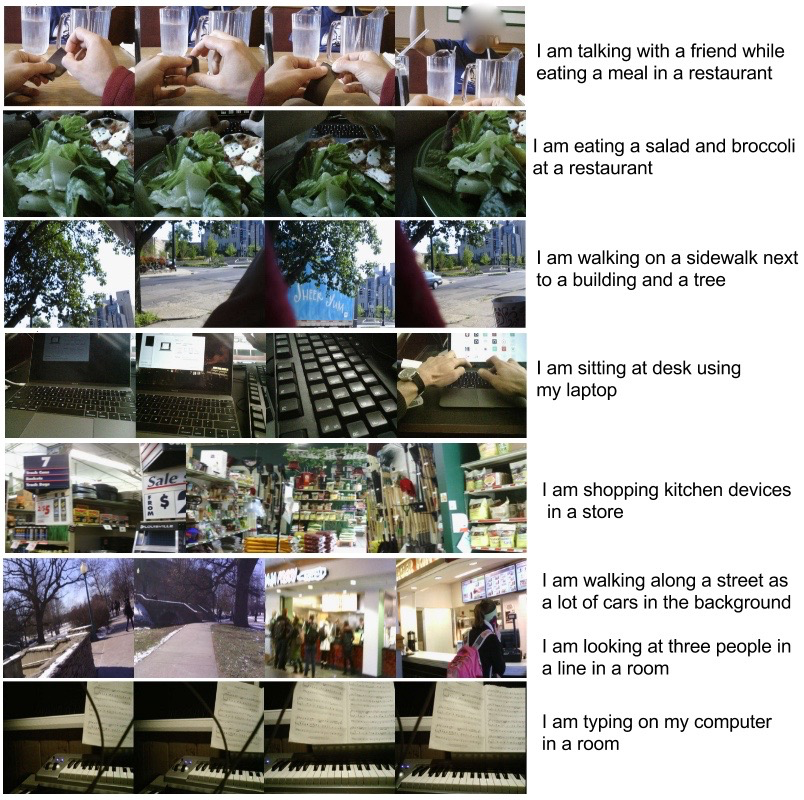}
\end{center}
\vspace{-6pt}
   \caption{Randomly-chosen samples of activity summarization on our  dataset.}
\label{fig:activity_caption}		
\end{figure*}

Finally, once captions have been jointly inferred for each
image in a photostream, we can group together contiguous substreams
of images that share the same sentence. 
Figure~\ref{fig:activity_caption} shows examples of activity
summarization. In general, the jointly-inferred captions are
reasonable descriptions of the images, and much less noisy than those
produced from individual images in Figure~\ref{fig:full_captions},
showing the advantage of incorporating temporal reasoning into the
captioning process.  For example, the first row of images shows that
the model labeled several images as ``I am talking with a friend while
eating a meal in a restaurant,'' even though the friend is only
visible in one of the frames, showing how the model has propagated
context across time. Of course, there are still mistakes
ranging from the minor error that there is no broccoli on the plate in
the second row to the more major error that the last row shows a piano
and not someone typing on a computer. The grammar of the sentences is
generally good considering that the model has no explicit knowledge of
English besides what it has learned from training data, although usage
errors are common (e.g., ``I am shopping kitchen devices in a store'').

\section{Experimental evaluation}
We first use automatic metrics that compare to ground truth reference sentences with quantitative scores. To give a better idea of the actual practical utility of technique, we also evaluate in two other ways: using a panel of human judges to rate the quality of captioning results, and testing the system in a specific application of keyword-based image retrieval using the generated captions.

\subsection{Quantitative captioning evaluation}

Automated metrics such as
BLEU~\cite{papineni2002bleu}, CIDEr~\cite{vedantam2015cider},
Meteor~\cite{banerjee2005meteor} and Rouge-L~\cite{lin2004rouge} have
been proposed to score sentence similarity compared to reference
sentences provided by humans, and each has different advantages and
disadvantages.  
BLEU was originally intended for evaluating machine
translations but is also commonly used to evaluate captioning.  BLEU
counts occurrences of n-grams in a candidate sentence (clipped
by
the maximum occurrences in the reference sentences), and normalizes by the
total number of n-grams in the candidate. It is typically evaluated
for different values of $n$, and in particular BLEU-$n$ refers to the
geometric mean from 1-grams to $n$-grams; here we report scores of
BLEU-1 to 4.  CIDEr~\cite{vedantam2015cider} computes average cosine
similarity (with TF-IDF weighting) between the $n$-grams (typically up
to 4-grams) of the generated sentence and human-generated reference
sentences.  Meteor~\cite{banerjee2005meteor} compares unigrams between
generated and reference sentences using different degrees of
similarity (exact match, matching stems, synonymy).
Rouge-L~\cite{lin2004rouge} produces an F-measure based on length of
longest common sequence of candidate and reference sequence.  
We present results using all of
these metrics (using the MS COCO Detection Challenge
implementation\footnote{https://github.com/tylin/coco-caption}), and
also summarize the seven scores with their mean.

\subsubsection{Implementation}

A significant challenge with deep learning-based methods is that
they typically require huge amounts of training data, both in terms of
images and sentences. Unfortunately, collecting this quantity of lifelogging
images and annotations is very difficult.  To try to
overcome this problem, we augmented our lifelogging training set with
COCO data using three different strategies: \textbf{Lifelog only}
training used only our lifelogging dataset, consisting of 736
lifelogging photos with 4,300 human-labeled sentences; \textbf{COCO
  only} training used only COCO dataset; and  \textbf{COCO then Lifelog} started with the
\textbf{COCO only} model, and then used it as initialization when
re-training the model on the lifelogging dataset 
(i.e.,  ``fine-tuning''~\cite{krizhevsky2012imagenet}).

For extracting image features, we use the
VGGNet~\cite{simonyan2014very} CNN model.  The word vectors are
learned from scratch.  Our image captioning model stacks two LSTM
layers, and each layer structure closely follows the one described
in~\cite{vinyals2014show}. To boost training speed, we re-implemented
LSTM model in C++ using the Caffe~\cite{jia2014caffe} deep
learning package. It
takes about 2.5 hours for COCO pre-training, and about 1 hour for
fine-tuning on Lifelog dataset with 10,000 iterations for both.


At test time, the number of beam searches conducted during caption
inference controls the degree of diversity in the output; here we use
three to match the three styles of captions we expect (COCO,
first-person, and third-person perspectives).  Samples of predicted
sentences are shown in Figure~\ref{fig:full_captions}. This suggests
that different genres of training sentences contribute to tune hidden
states of LSTM and thus enable it to produce diverse structures of
sentences in testing stage.


\begin{table*}[t] 
\centering
\ra{1.2}
{\small{\textsf{
\begin{tabular}{@{}llllrrrrrrrrr@{}}\toprule
&\multicolumn{2}{c}{\textit{Datasets}} & & \multicolumn{8}{c}{\textit{Metric}} & \phantom{i} \\
\cmidrule{2-3} \cmidrule{5-12} 
&Training &Testing & & Bleu-1 & Bleu-2  & Bleu-3 & Bleu-4 & CIDEr &METEOR& ROUGE & \,\,\,\, Mean\\ \midrule
&Lifelog& \multirow{3}{*}{Lifelog 100} && 0.669&0.472&0.324&0.218&0.257&0.209&0.462&0.373\\
&COCO&  && 0.561&0.354&0.206&0.118&0.143&0.149&0.374&0.272\\
&COCO+Lifelog & && 0.666&0.469&0.319&0.210&0.253&0.207&0.459&0.369 \\
&Lifelog@Usr1&Lifelog@Usr2&& 0.588&0.410&0.279&0.189&0.228&0.195&0.431&0.331 \\
&Lifelog@2015 & Lifelog@2016 &&0.557&0.379&0.249&0.160&0.325&0.202&0.425&0.328 \\
\bottomrule
\end{tabular}
}}}
\caption{Bleu1-4, CIDEr, Meteor and Rouge Scores for Diverse 3-Best Beams of Captions on Test Set. }
\label{tab:model_scores}
\end{table*}

\subsubsection{Results}
Table~\ref{tab:model_scores} presents quantitative results of each of
these training strategies, all tested on the same set of 100
randomly-selected photos having 1,000 ground truth reference
sentences, using each of the seven automatic scoring metrics mentioned
above.  We find that the \textbf{Lifelog only} strategy achieves much
higher overall accuracy than \textbf{COCO only}, with a mean score of
0.373 vs. 0.272. This suggests that even though COCO is a much larger
dataset, images from social media are different enough from
lifelogging images that the \textbf{COCO only} model does not
generalize well to our application. Moreover, this may also reflect an
artifact of the automated evaluation, because \textbf{Lifelog only}
benefits from seeing sentences with similar vocabulary and in a
similar style as in the reference sentences, since the same small
group of humans labeled both the training and test datasets.  More
surprisingly, we find that \textbf{Lifelog only} also slightly
outperforms \textbf{COCO then lifelog} (0.373 vs 0.369).
The model produced by the latter training dataset has a larger
vocabulary and produces richer styles of sentences than {Lifelog only},
which hurts its quantitative score.
Qualitatively, however, it often produces more diverse and descriptive sentences because
of its larger vocabulary and ability to generate 
sentences in first-person, third-person, and COCO styles.
Samples of generated diverse captions are shown in Figure~\ref{fig:full_captions}.


We conducted experiments with two additional strategies
in order to simulate more realistic scenarios.
The first scenario reflects when a consumer first starts using our
automatic captioning system on their images without having supplied
any training data of their own.  We simulate this by
training image captioning model on one user's photos and testing on
another. Training set has 805 photos and
3,716 reference sentences; testing set has 40 photos and 565
reference sentences.  The mean quantitative accuracy declines from our
earlier experiments when training and testing on images sampled from
the same set, as shown in Table~\ref{tab:model_scores}, although the
decline is not very dramatic (from 0.373 to 0.331), and still much
better than training on COCO (0.272). This result suggests that the
captioning model has learned general properties of lifelogging images,
instead of overfitting to one particular user (e.g., simply
``memorizing'' the appearance of the places and activities they
frequently visit and do).


The other situation is when
an existing model trained on historical lifelogging data is used to
caption new photos. We simulate this by taking all
lifelogging photos in 2015 as training data and photos in 2016 as
testing data. Training set has 673 photos and
3,610 sentences; testing set has 30 photos and 172
sentences. As shown in Table~\ref{tab:model_scores}, this
scenario very slightly decreased performance compared to training on data
from a different user (0.328 vs 0.331), although the difference is likely
not statistically significant.

\subsection{Evaluation with human judges}

The evaluation metrics used in the last section are convenient because
they can be automatically computed from ground-truth reference
sentences, and are helpful for objectively comparing different
methods. However, they give little insight into how accurate or
descriptive they are, or whether they would be useful for real
lifelogging users.
 

We conducted a small
study using human judges to rate the quality of our
automatically-generated captions. In particular, we randomly selected
21 images from the Lifelog 100 test dataset (used in
Table~\ref{tab:model_scores}) and generated captions using our model
trained on the {COCO then Lifelog} scenario. For each image, we
generated 15 captions (with 3 rounds of beam search, each with beam
size 5), and then kept the top-scoring caption according to our model
and four randomly-sampled from the remaining 14, to produce a diverse
set of five automatically-generated captions per image. We also randomly
sampled five of the human-generated reference sentences for each image.

For each of the ten captions (five automatic plus five human), we
showed the image (after reviewing it for potentially private content and
obtaining permission of the photo-taker) and caption to a user on Amazon Mechanical Turk,
without telling them how the caption had been produced.  We asked them
to rate, on a five-point Likert scale, how strongly they agreed with
two statements: (1) ``The sentence or phrase makes sense and is
grammatically correct (ignoring minor problems like capitalization and
punctuation,'' and (2) ``The sentence or phrase accurately describes
either what the camera wearer was doing or what he or she was looking
at when the photo was taken.'' 
The task involved 630 individual HITs from 37 users.

Table~\ref{tab:humans} summarizes the results, comparing the average
ratings over the 5 human reference sentences, the average over all 5
diverse automatically-generated captions (Auto-5 column), and the
single highest-likelihood caption as estimated by our complete model
(Auto-top).  About 92\% of the human reference sentences were judged
as grammatically correct (i.e., somewhat or strongly agreeing with
statement (1)), compared to about 77\% for the automatically-generated
diverse captions and 81\% for the single best sentence selected by our
model.  Humans also described images more accurately than the diverse
captions (88\% vs 54\%), although the fact that 64\% of our single
best estimated captions were accurate indicates that our model is
often able to identify which one is best among the diverse candidates.
Overall, our top automatic caption was judged to be \textit{both}
grammatically correct and accurate 59.5\% of the time, compared to
84.8\% of the time for human reference sentences. 

We view these results to be
very promising, as they suggest that automatic captioning can generate
reasonable sentences for over half of lifelogging images, at least in
some applications.  
For example, for 19 (90\%) of the
21 images in the test set, at least one of five diverse captions
was unanimously judged to be both grammatically correct and accurate
by all 3 judges. This may be useful in some retrieval
applications where recall is important, for example, where having
noise in some captions may be tolerable as long as at least one
of them is correct. We consider one such application in the next section.

\begin{table*}[t]\centering
\ra{1.0}
{\small{\textsf{
\begin{tabular}{@{\,}rrrr@{\,}crrr@{\,}}\toprule
& \multicolumn{3}{c}{\textit{Grammar}} && \multicolumn{3}{c}{\textit{Accuracy}} \\
\cmidrule{2-4} \cmidrule{6-8} 
\textit{Rating} & Human & Auto-5 & Auto-top& & Human & Auto-5 & Auto-top \\ \midrule
1 & 1.9\% & 7.6\% & 11.9\%&& 2.9\% & 22.4\% & 21.4\% \\
2 & 3.8\% & 10.0\% & 7.1\%&& 3.8\% & 15.2\% & 7.1\% \\
3 & 0.5\% & 5.7\% & 0.0\%&& 4.8\% & 8.1\% & 7.1\% \\
4 & 19.0\% & 17.6\% & 4.8\%&& 22.4\% & 17.6\% & 19.0\% \\
5 & 73.3\% & 59.0\% & 76.2\%&& 65.2\% & 36.7\% & 45.2\% \\ \midrule
Mean & 4.60 & 4.10 & 4.26 && 4.45 & 3.31 & 3.60 \\ \bottomrule
\end{tabular}
}}}
\caption{Summary of  grammatical correctness
  and accuracy of lifelogging image captions, on a rating scale from 1
  (Strongly Disagree) to 5 (Strongly Agree), averaged over 3
  judges. \textit{Human} column is averaged over 5 human-generated
  reference sentences, \textit{Auto-5} is averaged over 5 diverse
  computer-generated sentences, and \textit{Auto-top} is single highest-likelihood
  computer-generated sentence predicted by our model.  }
\label{tab:humans}
\end{table*}

\subsection{Keyword-based image retrieval}

Image captioning allows us to directly implement keyword-based image retrieval by
searching on the generated captions.  We consider a particular application 
of this image search feature here
that permits a quantitative evaluation.  As mentioned above, wearable
cameras can collect a large number of images containing private
information. Automatic image captioning could allow users to find
potentially private images easily, and then take appropriate action
(like deleting or encrypting the photos).  We consider two specific
types of potentially embarrassing content here: photos taken in
potentially private locations like bathrooms and locker rooms, and
photos containing personal computer or smartphone displays which may
contain private information such as credit card numbers or e-mail
contents. 

We chose these two types of concerns specifically because they have
been considered by others in prior work: Korayem
\etal~\cite{objectavoider} present a system for detecting monitors in
lifelogging images using deep learning with CNNs, while Templeman
\etal~\cite{templeman2014placeavoider} classify images according to the room in
which they were taken. Both of these papers present strongly
supervised based techniques, which were given thousands of training
images manually labeled with ground truth for each particular task.
In contrast, identifying private imagery based on keyword search on
automatically-generated captions could avoid the need to create a
training set and train a separate classifier for each type of sensitive
image.

We evaluated captioning-based sensitive image retrieval against
standard state-of-the-art strongly-supervised image classification
using CNNs~\cite{krizhevsky2012imagenet} (although we cannot compare
directly to the results presented in~\cite{objectavoider} or
\cite{templeman2014placeavoider} because we use different datasets).  We trained
the strongly-supervised model by first generating a training set
consisting of photos having monitors and not having monitors, and
photos taken in bathrooms and locker rooms or elsewhere, by using the
ground truth categories given in the COCO and Flickr8k datasets. This
yielded 34,736 non-sensitive images, 6,135 images taken in sensitive
places, and 4,379 images with displays. 
We used pre-trained AlexNet model (1000-way classifier on ImageNet data) and fine-tuned on
our dataset by replacing the final fully connected layer with a 3-way classifier
to correspond with our three-class problem.

\begin{table*}[t]
\vspace{4pt}
\begin{minipage}{0.53\textwidth} 
\centering
{\small{\ra{1.0}
{\textsf{
{\textbf{3-way classification}} \\
\begin{tabular}{@{}rrrrcrrrc@{}}\toprule
& \multicolumn{3}{c}{\textit{CNN-based}} & \,& \multicolumn{3}{c}{\textit{Caption-based}} &  \\
\cmidrule{2-4} \cmidrule{6-8} 
& NotSen & Place  & Display  && NotSen & Place  & Display \\ \midrule
NotSen &    0.730          &  0.130      & 0.140  & &    0.686          & 0.117            &  0.197   \\ 
Place     &     0.189         &  0.811        &   0 & &     0.151         &     0.792       &      0.057\\  
Display       &      0.300        &    0.043        & 0.657 & &    0.143            &      0.008      &    0.849 \\ 
\bottomrule
\end{tabular}
}}}}
\end{minipage}
\begin{minipage}{0.5\textwidth}
\centering
{\small{\ra{1.0}
{\textsf{
{\textbf{2-way classification}} \\
\begin{tabular}{@{}rrrrrr@{}}\toprule
& \multicolumn{2}{c}{\textit{CNN-based}} & \,& \multicolumn{2}{c}{\textit{Caption-based}}  \\
\cmidrule{2-3} \cmidrule{5-6} 
& NotSen &{Sen}&& NotSen& {Sen}\\ \midrule
NotSen    & 0.730 & 0.270 && 0.686 & 0.314  \\
Sen	    & 0.317& 0.683 && 0.161& 0.839 \\
\bottomrule \\
\end{tabular}
}}}}
\end{minipage}
\vspace{-6pt}
\caption{Confusion matrices for two approaches on two tasks for detecting sensitive images. \textit{Left:} Results on
3-way problem of classifying into not sensitive, sensitive place (bathroom), or digital display categories.
\textit{Right:} Results on 2-way problem of classifying into sensitive or not (regardless of sensitivity type).
Actual classes are in rows and predicted classes are in columns.}
\label{tab:confuse_sensitive_1}
\end{table*}

We also ran the technique proposed here, where we first generate
automatic image captions, and then search through the top five
captions for each image for a set of pre-defined keywords
(specifically ``toilet,'' ``bathroom,'' ``locker,'' ``lavatory,'' and
``washroom'' for sensitive place detection, and ``computer,''
``laptop,'' ``iphone,'' ``smartphone,'' and ``screen'' for display
detection). If any of these keywords is detected in any of the five
captions, the image is classified as sensitive, and otherwise it is estimated to be not sensitive.

\begin{figure}[t]
  \centering
  \begin{tabular}{@{}c@{}c@{}}
\includegraphics[width=0.4\columnwidth,trim=1.2cm 0cm 1.7cm 1.5cm,clip]{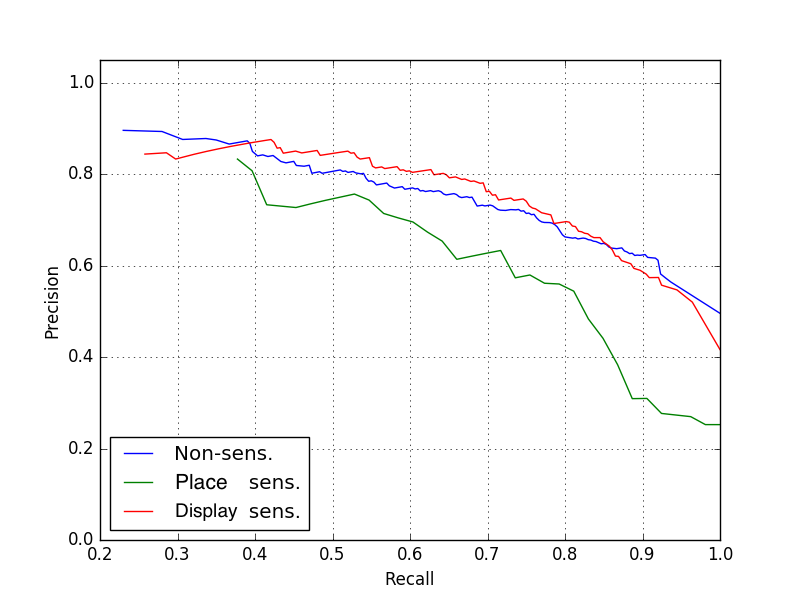} &
\includegraphics[width=0.4\columnwidth,trim=1.2cm 0cm 1.7cm 1.5cm,clip]{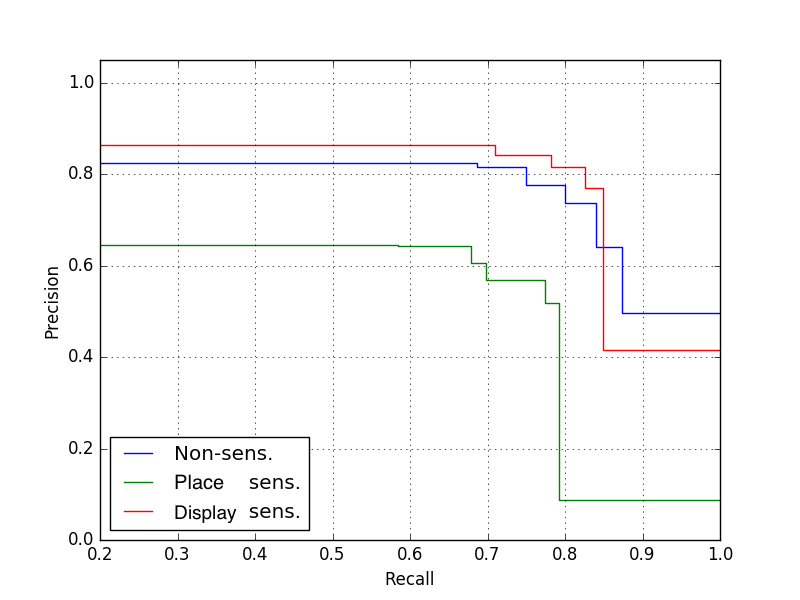} \\
\end{tabular}
\vspace{-12pt}
  \caption{Precision-recall curves for retrieving sensitive images using CNNs (left) and generated captions (right).}
  \label{fig:pr_curves}		
\end{figure}

Table~\ref{tab:confuse_sensitive_1} presents the confusion matrix for
each method, using a set of 600 manually-annotated images from our
lifelogging dataset as test data (with 300 non-sensitive images, 53
images in sensitive places, and 252 with digital displays). We see the
supervised classifier has better prediction performance on finding
sensitive places (0.811) than keyword based classifiers (0.792), while
the caption-based technique classifier outperforms on predicting
second type of sensitive images (0.849 vs 0.657). In a real application,
determining the type of private image is likely less important 
than simply deciding if it  is private. The right table in
Table~\ref{tab:confuse_sensitive_1} reflects this scenario, showing a confusion matrix which
combines the two sensitive types and focuses on whether photos are
sensitive or not. 

From another point of view, sensitive photo detection 
is a retrieval problem.  Figure~\ref{fig:pr_curves} shows
precision-recall curves for CNN and caption-based classifiers,
respectively.  They show the trade-off between selecting accurate
sensitive photos (high precision) and obtaining a majority of all
sensitive photos (high recall). For example, by using CNN classifier,
we can obtain 80\% type 1 (sensitive place) photos with accuracy around
58\% (Figure~\ref{fig:pr_curves}(left) green curve); by using
the caption-based classifier, we can obtain 80\% of type 2 (digital display)
sensitive photos with precision around
78\% (Figure~\ref{fig:pr_curves}(right) blue curve).

Overall, these results suggest that keyword search in
automatically-generated captions could yield similar accuracies to 
strongly-supervised classifiers, but without
having to be explicitly re-trained on each
type of private image.
The two approaches may also be complementary, since they use
different forms of evidence in making classification decisions, and
users in a real application could choose their own trade-off on how
aggressively to filter lifelogging images.

\section{Conclusion}
In this paper, we have proposed the concept of using automatically-generated
captions to help organize and annotate lifelogging image collections.
We have proposed a deep learning-based captioning model that
jointly labels photo streams in order to take advantage of temporal consistency between photos.
Our evaluation suggests that modern automated
captioning techniques could work well enough to be used in practical 
lifelogging photo applications.
We hope our research will motivate further efforts of using lifelogging photos and descriptions together to help human memory recall the activities and scenarios. 

\section{Acknowledgements}
This work was supported in part by the National Science Foundation
(IIS-1253549 and CNS-1408730) and Google, and used compute facilities
provided by NVidia, the Lilly Endowment through support of the IU PTI, and the Indiana METACyt Initiative. We
thank Zhenhua Chen, Sally Crandall, and Xuan Dong for helping to label
our lifelogging photos.

\bibliographystyle{ieee}
\bibliography{deepdiary_arXiv}

\begin{thebibliography}{10}\itemsep=-1pt

\bibitem{azuma2001recent}
R.~Azuma, Y.~Baillot, R.~Behringer, S.~Feiner, S.~Julier, and B.~MacIntyre.
\newblock Recent advances in augmented reality.
\newblock {\em IEEE Computer Graphics and Applications}, 21(6):34--47, 2001.

\bibitem{egohands2015iccv}
S.~Bambach, S.~Lee, D.~Crandall, and C.~Yu.
\newblock Lending a hand: Detecting hands and recognizing activities in complex
  egocentric interactions.
\newblock In {\em IEEE Intl. Conf. on Computer Vision}, 2015.

\bibitem{banerjee2005meteor}
S.~Banerjee and A.~Lavie.
\newblock {METEOR}: An automatic metric for {MT} evaluation with improved
  correlation with human judgments.
\newblock {\em ACL Workshop on Intrinsic and Extrinsic Evaluation Measures for
  Machine Translation and/or Summarization}, pages 65--72, 2005.

\bibitem{barreau2006augmenting}
D.~Barreau, A.~Crystal, J.~Greenberg, A.~Sharma, M.~Conway, J.~Oberlin,
  M.~Shoffner, and S.~Seiberling.
\newblock Augmenting memory for student learning: Designing a context-aware
  capture system for biology education.
\newblock {\em American Society for Information Science and Technology},
  43(1):1--6, 2006.

\bibitem{batra2012diverse}
D.~Batra, P.~Yadollahpour, A.~Guzman-Rivera, and G.~Shakhnarovich.
\newblock Diverse m-best solutions in markov random fields.
\newblock In {\em European Conf. on Computer Vision}, pages 1--16. Springer,
  2012.

\bibitem{castro2015activities}
D.~Castro, S.~Hickson, V.~Bettadapura, E.~Thomaz, G.~Abowd, H.~Christensen, and
  I.~Essa.
\newblock Predicting daily activities from egocentric images using deep
  learning.
\newblock In {\em Intl. Symposium on Wearable Computers}, 2015.

\bibitem{clinch2014lifelogging}
S.~Clinch, P.~Metzger, and N.~Davies.
\newblock Lifelogging for observer view memories: an infrastructure approach.
\newblock In {\em 2014 ACM Intl. Joint Conf. on Pervasive and Ubiquitous
  Computing: Adjunct Publication}, pages 1397--1404, 2014.

\bibitem{deepdiary2016eccvw}
D.~Crandall and C.~Fan.
\newblock Deepdiary: Automatically captioning lifelogging image streams.
\newblock In {\em European Conference on Computer Vision International Workshop
  on Egocentric Perception, Interaction, and Computing}.

\bibitem{dalal2005histograms}
N.~Dalal and B.~Triggs.
\newblock Histograms of oriented gradients for human detection.
\newblock In {\em IEEE Conf. on Computer Vision and Pattern Recognition},
  volume~1, pages 886--893, 2005.

\bibitem{denning2014situ}
T.~Denning, Z.~Dehlawi, and T.~Kohno.
\newblock In situ with bystanders of augmented reality glasses: Perspectives on
  recording and privacy-mediating technologies.
\newblock In {\em ACM SIGCHI Conf. on Human Factors in Computing Systems},
  pages 2377--2386, 2014.

\bibitem{doherty2012experiences}
A.~Doherty, K.~Pauly-Takacs, N.~Caprani, C.~Gurrin, C.~Moulin, N.~O'Connor, and
  A.~Smeaton.
\newblock Experiences of aiding autobiographical memory using the sensecam.
\newblock {\em Human--Computer Interaction}, 27(1-2):151--174, 2012.

\bibitem{doherty2011passively}
A.~R. Doherty, N.~Caprani, V.~Kalnikaite, C.~Gurrin, A.~F. Smeaton, E.~Noel,
  et~al.
\newblock Passively recognising human activities through lifelogging.
\newblock {\em Comput. Hum. Behav.}, 27(5):1948--1958, 2011.

\bibitem{elman1990finding}
J.~L. Elman.
\newblock Finding structure in time.
\newblock {\em Cognitive science}, 14(2):179--211, 1990.

\bibitem{erhan2014scalable}
D.~Erhan, C.~Szegedy, A.~Toshev, and D.~Anguelov.
\newblock Scalable object detection using deep neural networks.
\newblock In {\em IEEE Conf. on Computer Vision and Pattern Recognition}, pages
  2155--2162, 2014.

\bibitem{fathi2012learning}
A.~Fathi, Y.~Li, and J.~M. Rehg.
\newblock Learning to recognize daily actions using gaze.
\newblock In {\em European Conf. on Computer Vision}, pages 314--327. Springer,
  2012.

\bibitem{fathi2011learning}
A.~Fathi, X.~Ren, and J.~M. Rehg.
\newblock Learning to recognize objects in egocentric activities.
\newblock In {\em IEEE Conf. on Computer Vision and Pattern Recognition}, pages
  3281--3288, 2011.

\bibitem{furnari2015}
A.~Furnari, G.~Farinella, and S.~Battiano.
\newblock Recognizing personal contexts from egocentric images.
\newblock In {\em ICCV Workshops}, 2015.

\bibitem{girshick2014rich}
R.~Girshick, J.~Donahue, T.~Darrell, and J.~Malik.
\newblock Rich feature hierarchies for accurate object detection and semantic
  segmentation.
\newblock In {\em IEEE Conf. on Computer Vision and Pattern Recognition}, pages
  580--587, 2014.

\bibitem{graves2013generating}
A.~Graves.
\newblock Generating sequences with recurrent neural networks.
\newblock {\em arXiv:1308.0850}, 2013.

\bibitem{graves2013hybrid}
A.~Graves, N.~Jaitly, and A.-R. Mohamed.
\newblock Hybrid speech recognition with deep bidirectional lstm.
\newblock In {\em IEEE Workshop on Automatic Speech Recognition and
  Understanding}, pages 273--278, 2013.

\bibitem{gurrin2008examination}
C.~Gurrin, A.~F. Smeaton, D.~Byrne, N.~Hare, G.~J. Jones, and N.~Connor.
\newblock An examination of a large visual lifelog.
\newblock In {\em Information Retrieval Technology}, pages 537--542. Springer,
  2008.

\bibitem{hochreiter1997long}
S.~Hochreiter and J.~Schmidhuber.
\newblock Long short-term memory.
\newblock {\em Neural computation}, 9(8):1735--1780, 1997.

\bibitem{hodges2006sensecam}
S.~Hodges, L.~Williams, E.~Berry, S.~Izadi, J.~Srinivasan, A.~Butler, G.~Smyth,
  N.~Kapur, and K.~Wood.
\newblock Sensecam: A retrospective memory aid.
\newblock In {\em ACM Conf. on Ubiquitous Computing}, pages 177--193, 2006.

\bibitem{hoyle2014privacy}
R.~Hoyle, R.~Templeman, S.~Armes, D.~Anthony, D.~Crandall, and A.~Kapadia.
\newblock Privacy behaviors of lifeloggers using wearable cameras.
\newblock In {\em ACM Intl. Joint Conf. on Pervasive and Ubiquitous Computing},
  pages 571--582, 2014.

\bibitem{jia2014caffe}
Y.~Jia, E.~Shelhamer, J.~Donahue, S.~Karayev, J.~Long, R.~Girshick,
  S.~Guadarrama, and T.~Darrell.
\newblock Caffe: Convolutional architecture for fast feature embedding.
\newblock {\em arXiv:1408.5093}, 2014.

\bibitem{kalnikaite2010now}
V.~Kalnikaite, A.~Sellen, S.~Whittaker, and D.~Kirk.
\newblock Now let me see where {I} was: Understanding how lifelogs mediate
  memory.
\newblock In {\em ACM SIGCHI Conf. on Human Factors in Computing Systems},
  pages 2045--2054, 2010.

\bibitem{karim2006exploiting}
S.~Karim, A.~Andjomshoaa, and A.~Tjoa.
\newblock Exploiting sensecam for helping the blind in business negotiations.
\newblock In {\em Computers Helping People with Special Needs}. Springer, 2006.

\bibitem{karpathy2014deep}
A.~Karpathy and L.~Fei-Fei.
\newblock Deep visual-semantic alignments for generating image descriptions.
\newblock {\em arXiv:1412.2306}, 2014.

\bibitem{karpathy2015visualizing}
A.~Karpathy, J.~Johnson, and L.~Fei-Fei.
\newblock Visualizing and understanding recurrent networks.
\newblock {\em arXiv:1506.02078}, 2015.

\bibitem{karpathy2014fragment}
A.~Karpathy, A.~Joulin, and L.~Fei-Fei.
\newblock Deep fragment embeddings for bidirectional image sentence mapping.
\newblock In {\em Advances in neural information processing systems}, pages
  1889--1897, 2014.

\bibitem{koller}
D.~Koller and N.~Friedman.
\newblock {\em Probabilistic Graphical Models: Principles and Techniques}.
\newblock MIT Press, 2009.

\bibitem{objectavoider}
M.~Korayem, R.~Templeman, D.~Chen, D.~Crandall, and A.~Kapadia.
\newblock Enhancing lifelogging privacy by detecting screens.
\newblock In {\em ACM CHI Conf. on Human Factors in Computing Systems}, 2016.

\bibitem{krizhevsky2012imagenet}
A.~Krizhevsky, I.~Sutskever, and G.~Hinton.
\newblock Imagenet classification with deep convolutional neural networks.
\newblock In {\em Advances in Neural Information Processing Systems}, pages
  1097--1105, 2012.

\bibitem{lin2004rouge}
C.-Y. Lin.
\newblock Rouge: A package for automatic evaluation of summaries.
\newblock In {\em Workshop On Text Summarization Branches Out}, 2004.

\bibitem{mscoco}
T.-Y. Lin, M.~Maire, S.~Belongie, J.~Hays, P.~Perona, D.~Ramanan,
  P.~Doll{\'a}r, and C.~L. Zitnick.
\newblock Microsoft {COCO}: Common objects in context.
\newblock In {\em European Conf. on Computer Vision}, pages 740--755. Springer,
  2014.

\bibitem{lowe2004distinctive}
D.~Lowe.
\newblock Distinctive image features from scale-invariant keypoints.
\newblock {\em Intl. J. of computer vision}, 60(2):91--110, 2004.

\bibitem{mann2002sousveillance}
S.~Mann, J.~Nolan, and B.~Wellman.
\newblock Sousveillance: Inventing and using wearable computing devices for
  data collection in surveillance environments.
\newblock {\em Surveillance \& Society}, 1(3):331--355, 2002.

\bibitem{mao2014explain}
J.~Mao, W.~Xu, Y.~Yang, J.~Wang, and A.~L. Yuille.
\newblock Explain images with multimodal recurrent neural networks.
\newblock {\em arXiv:1410.1090}, 2014.

\bibitem{policecams}
L.~Miller and J.~Toliver.
\newblock Implementing a body-worn camera program: Recommendations and lessons
  learned.
\newblock Technical report, Office of Community Oriented Policing Services,
  2014.

\bibitem{moghimi2014sedentary}
M.~Moghimi, W.~Wu, J.~Chen, S.~Godbole, S.~Marshall, J.~Kerr, and S.~Belongie.
\newblock Analyzing sedentary behavior in life-logging images.
\newblock In {\em Intl. Conf. on Image Processing}, 2014.

\bibitem{nguyen2009encountering}
D.~H. Nguyen, G.~Marcu, G.~R. Hayes, K.~N. Truong, J.~Scott, M.~Langheinrich,
  and C.~Roduner.
\newblock Encountering sensecam: personal recording technologies in everyday
  life.
\newblock In {\em ACM Intl. Conf. on Ubiquitous Computing}, pages 165--174,
  2009.

\bibitem{papineni2002bleu}
K.~Papineni, S.~Roukos, T.~Ward, and W.-J. Zhu.
\newblock {BLEU: a method for automatic evaluation of machine translation}.
\newblock In {\em Annual Meeting of the Association for Computational
  Linguistics}, pages 311--318, 2002.

\bibitem{razavian2014cnn}
A.~S. Razavian, H.~Azizpour, J.~Sullivan, and S.~Carlsson.
\newblock Cnn features off-the-shelf: an astounding baseline for recognition.
\newblock In {\em IEEE Conf. on Computer Vision and Pattern Recognition
  Workshops}, pages 512--519, 2014.

\bibitem{ryoo2015robot}
M.~Ryoo, T.~J. Fuchs, L.~Xia, J.~K. Aggarwal, and L.~Matthies.
\newblock Robot-centric activity prediction from first-person videos: What will
  they do to me.
\newblock In {\em ACM/IEEE Intl. Conf. on Human-Robot Interaction}, pages
  295--302, 2015.

\bibitem{ryoo2013first}
M.~Ryoo and L.~Matthies.
\newblock First-person activity recognition: What are they doing to me?
\newblock In {\em IEEE Conf. on Computer Vision and Pattern Recognition}, pages
  2730--2737, 2013.

\bibitem{schuster1997bidirectional}
M.~Schuster and K.~Paliwal.
\newblock Bidirectional recurrent neural networks.
\newblock {\em IEEE T. Signal Processing}, 45(11):2673--2681, 1997.

\bibitem{simonyan2014very}
K.~Simonyan and A.~Zisserman.
\newblock Very deep convolutional networks for large-scale image recognition.
\newblock {\em arXiv:1409.1556}, 2014.

\bibitem{szegedy2013deep}
C.~Szegedy, A.~Toshev, and D.~Erhan.
\newblock Deep neural networks for object detection.
\newblock In {\em Advances in Neural Information Processing Systems}, pages
  2553--2561, 2013.

\bibitem{templeman2014placeavoider}
R.~Templeman, M.~Korayem, D.~J. Crandall, and A.~Kapadia.
\newblock Placeavoider: Steering first-person cameras away from sensitive
  spaces.
\newblock In {\em Network and Distributed Systems Security Symposium}, 2014.

\bibitem{vedantam2015cider}
R.~Vedantam, C.~Zitnick, and D.~Parikh.
\newblock Cider: Consensus-based image description evaluation.
\newblock In {\em IEEE Conf. on Computer Vision and Pattern Recognition}, pages
  4566--4575, 2015.

\bibitem{venugopalan2015sequence}
S.~Venugopalan, M.~Rohrbach, J.~Donahue, R.~Mooney, T.~Darrell, and K.~Saenko.
\newblock Sequence to sequence--video to text.
\newblock {\em arXiv:1505.00487}, 2015.

\bibitem{vinyals2014show}
O.~Vinyals, A.~Toshev, S.~Bengio, and D.~Erhan.
\newblock Show and tell: A neural image caption generator.
\newblock {\em arXiv:1411.4555}, 2014.

\bibitem{yoo2013facelog}
C.~Yoo, J.~Shin, I.~Hwang, and J.~Song.
\newblock Facelog: capturing user's everyday face using mobile devices.
\newblock In {\em ACM Conf. on Pervasive and Ubiquitous Computing}, pages
  163--166, 2013.

\bibitem{zhu2015aligning}
Y.~Zhu, R.~Kiros, R.~Zemel, R.~Salakhutdinov, R.~Urtasun, A.~Torralba, and
  S.~Fidler.
\newblock Aligning books and movies: Towards story-like visual explanations by
  watching movies and reading books.
\newblock {\em arXiv:1506.06724}, 2015.

\end{thebibliography}
\end{document}